\documentclass[sigconf]{acmart}

\usepackage{booktabs} 
\usepackage{multirow}
\usepackage{color}
\usepackage{amsmath}
\usepackage{balance} 

\AtBeginDocument{%
  \providecommand\BibTeX{{%
    \normalfont B\kern-0.5em{\scshape i\kern-0.25em b}\kern-0.8em\TeX}}}

\copyrightyear{2021} 
\acmYear{2021} 
\setcopyright{acmcopyright}\acmConference[SIGIR '21]{Proceedings of the 44rd International ACM SIGIR Conference on Research and Development in Information Retrieval}{July 11-15, 2021}{}
\acmBooktitle{Proceedings of the 44rd International ACM SIGIR Conference on Research and Development in Information Retrieval (SIGIR '21), July 11-15, 2021}

\begin{document}
\fancyhead{}

\title{Hierarchically Modeling Micro and Macro Behaviors via Multi-Task Learning for Conversion Rate Prediction}



\author{Hong Wen}
\authornote{Co-first authors contributed equally to this paper.}
\affiliation{%
  \institution{Alibaba Group}
  \city{Hangzhou, Zhejiang, China 311121}
}
\email{qinggan.wh@alibaba-inc.com}

\author{Jing Zhang}
\authornotemark[1]
\affiliation{%
  \institution{The University of Sydney}
  \city{Darlington NSW 2008, Australia}
}
\email{jing.zhang1@sydney.edu.au}

\author{Fuyu Lv}
\affiliation{%
  \institution{Alibaba Group}
  \city{Hangzhou, Zhejiang, China 311121}
}
\email{fuyu.lfy@alibaba-inc.com}

\author{Wentian Bao}
\affiliation{%
  \institution{Alibaba Group}
  \city{Hangzhou, Zhejiang, China 311121}
}
\email{wentian.bwt@alibaba-inc.com}

\author{Tianyi Wang}
\affiliation{%
  \institution{Alibaba Group}
  \city{Hangzhou, Zhejiang, China 311121}
}
\email{joshua.wty@alibaba-inc.com}

\author{Zulong Chen}
\affiliation{%
  \institution{Alibaba Group}
  \city{Hangzhou, Zhejiang, China 311121}
}
\email{zulong.czl@alibaba-inc.com}


\begin{abstract}


Conversion Rate (\emph{CVR}) prediction in modern industrial e-commerce platforms is becoming increasingly important, which directly contributes to the final revenue. 
In order to address the well-known sample selection bias (\emph{SSB}) and data sparsity (\emph{DS}) issues encountered during CVR modeling, the abundant labeled macro behaviors ($i.e.$, user's interactions with items) are used. Nonetheless, we observe that several purchase-related micro behaviors ($i.e.$, user's interactions with specific components on the item detail page) can supplement fine-grained cues for \emph{CVR} prediction. Motivated by this observation, we propose a novel \emph{CVR} prediction method by Hierarchically Modeling both Micro and Macro behaviors ($HM^3$). Specifically, we first construct a complete user sequential behavior graph to hierarchically represent micro behaviors and macro behaviors as one-hop and two-hop post-click nodes. Then, we embody $HM^3$ as a multi-head deep neural network, which predicts six probability variables corresponding to explicit sub-paths in the graph. They are further combined into the prediction targets of four auxiliary tasks as well as the final $CVR$ according to the conditional probability rule defined on the graph. By employing multi-task learning and leveraging the abundant supervisory labels from micro and macro behaviors, $HM^3$ can be trained end-to-end and address the \emph{SSB} and \emph{DS} issues. Extensive experiments on both offline and online settings demonstrate the superiority of the proposed $HM^3$ over representative state-of-the-art methods.


\end{abstract}




\maketitle

\setlength{\abovecaptionskip}{0.2cm}
\setlength{\belowcaptionskip}{-0.4cm}

\section{Introduction}
E-commerce Recommender Systems (\emph{RS}) \cite{schafer1999recommender,wei2007survey,zhang2019deep,huseynov2020intelligent,gong2020edgerec} serve a vital role to help users discover their preferred items from a vast number of candidates, therefore improving the user experience \cite{xu2015customer} and delivering new business value \cite{lu2014show,gong2020edgerec,zhang2020empowering}. A typical \emph{RS} usually consists of two significant phases, $i.e.$, retrieval and ranking, where Click-Through Rate (\emph{CTR}) \cite{guo2017deepfm,zhou2018deep,zhou2019deep, feng2019deep,pi2019practice} and Conversion Rate (\emph{CVR}) \cite{yang2016large,lu2017practical,ma2018entire,wen2019multi,wen2020entire} predictions in the latter phase are two fundamental tasks, measuring the probability of an item from impression to click and from click to purchase, respectively. In this paper, we focus on the post-click \emph{CVR} estimation task. 

In general, conventional \emph{CVR} modeling methods always employ similar techniques developed for \emph{CTR} tasks, where the only difference is replacing the training samples generated from the path ``impression$\to$click'' to ``click$\to$purchase'' in the user sequential behavior graph \cite{ma2018entire,wen2020entire}. Although it is simple and effective for online development, two critical issues encountered in practice, $i.e.$, sample selection bias (\emph{SSB}) \cite{zadrozny2004learning} and data sparsity (\emph{DS}) \cite{lee2012estimating}. \emph{SSB} refers to the issue that the training and inference sample space of CVR prediction can be systematically different, and \emph{DS} refers to the fact the number of training samples for CVR prediction is much less than that for CTR prediction. These two issues affect industrial-level recommender systems. Furthermore, the deficiency of positive samples also increases the difficulty of \emph{CVR} modeling, $e.g.$, only less than 0.1\% of impressions will finally convert to purchases, according to the statistics from our e-commerce platform.

\begin{figure*}
  \centering
  \includegraphics[width=0.82\linewidth]{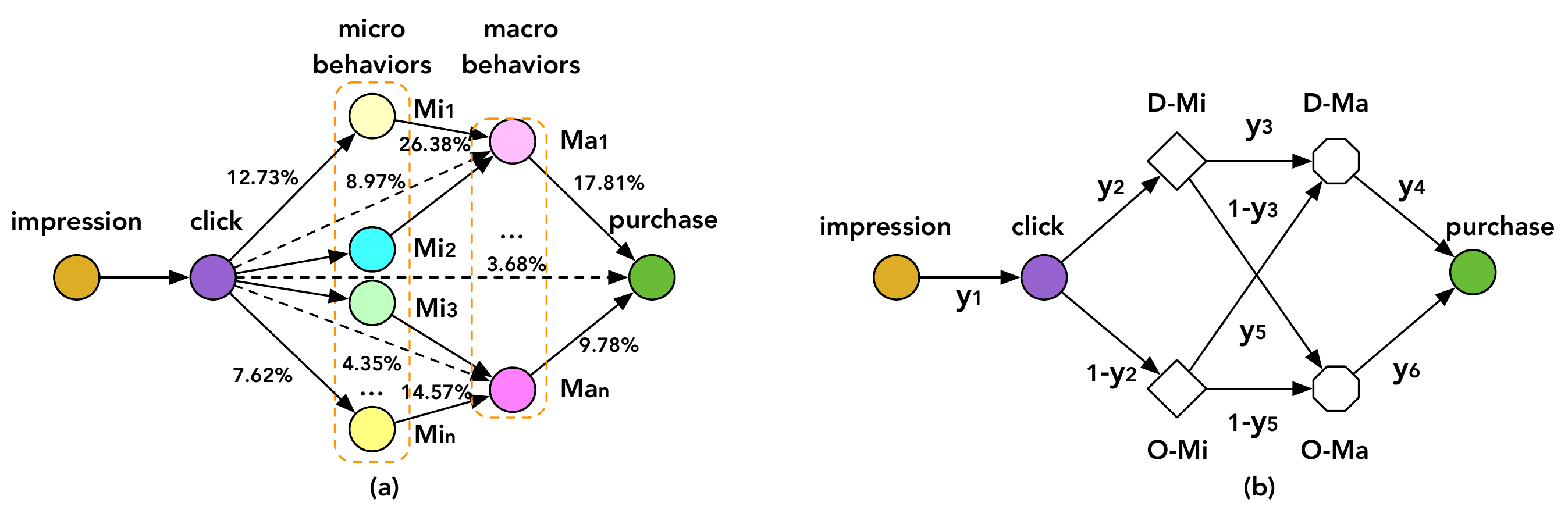}
  \caption{Illustration of the motivation of our method. (a) Hierarchically inserting micro and macro behaviors between click and purchase. Here, $M_{i_{1}}$,$M_{i_{2}}$,..$M_{i_{n}}$ represent the micro behaviors while $M_{a_{1}}$,..$M_{a_{n}}$ represent the macro behaviors (See Table~\ref{table:mm_behaviors}). The number above each edge represents the sparsity of respective path. (b) Dividing micro (macro) behaviors into two disjoint sets namely \emph{D-Mi} and \emph{O-Mi} ( \emph{D-Ma} and \emph{O-Ma}) results in a novel user sequential behavior graph.}
  \label{fig:HM3_motivation}
\end{figure*}


To address these challenges, researchers try to model \emph{CVR} prediction and learn strong feature representations \cite{athiwaratkun2015feature, huang2019multimodal} over the entire space with all impression samples. For example, 
\emph{ESMM} \cite{ma2018entire} models the ``impression$\to$click$\to$purchase'' path for the CVR task, and trains the CVR model over the entire space. In this way, ESMM can effectively mitigate the \emph{SSB} and \emph{DS} issues and achieves better performance than several conventional \emph{CVR} modeling methods. However, it still faces the challenge that the positive samples of \emph{CVR} prediction are insufficient. In fact, users always take many purchase-related behaviors after clicking an item, $e.g.$, putting it to the shopping cart instead of immediately purchasing it. Motivated by this observation, $ESM^{2}$ \cite{wen2020entire} models \emph{CVR} prediction and some auxiliary tasks simultaneously in a multi-task learning framework based on the conditional probability rule elaborately defined on a new user behavior graph. However, these post-click behaviors in $ESM^{2}$ are all macro behaviors, which can only provide understandings of the subsequent purchase behavior at the item-level granularity, while more abundant fine-grained interactions are ignored, $e.g.$, clicks on certain components of items like comments or pictures, which we call micro behaviors in this paper. Micro behaviors can provide more detailed supplemental understandings of the subsequent macro behaviors and purchase behavior.

Few existing works pay attention to these valuable micro behaviors. To the best of our knowledge, \citeauthor{zhou2018micro}~\cite{zhou2018micro} firstly proposed to utilize user micro-behaviors in e-commerce recommendation, offering a new perspective to understand user behavior modeling. \citeauthor{gu2020hierarchical}~\cite{gu2020hierarchical} proposed a hierarchical user profiling framework to solve the user profiling problem in e-commerce recommender systems, which not only utilizes macro behaviors but also harvests users’ fine-grained micro behaviors. For \emph{CVR} prediction, \citeauthor{bao2020gmcm}~\cite{bao2020gmcm} proposed to utilize the micro behaviors to provide explicit instructions for \emph{CVR} modeling. Specifically, it represents the micro behaviors as a graph and employs graph convolutional networks to thoroughly model the relationships among all the micro behaviors towards the final purchase target. However, it ignores the macro behaviors, which have been demonstrated their value in \cite{wen2020entire}. How to explicitly model both micro and macro behaviors in a unified framework for \emph{CVR} prediction remains challenging and unexplored. 

In this paper, we fill this gap by proposing a novel $CVR$ prediction model named $HM^3$ which can hierarchically model both micro and macro behaviors in a unified framework. First, we construct a complete user sequential behavior graph to hierarchically represent micro behaviors and macro behaviors as one-hop and two-hop post-click nodes, respectively. Then, we embody $HM^3$ as a multi-head deep neural network (DNN) like \cite{wen2020entire}. Specifically, $HM^3$ predicts six probability variables corresponding to explicit sub-paths in the graph from six parallel sub-networks after a shared feature embedding module. They are further combined into the targets of four auxiliary tasks and the final \emph{CVR} according to the conditional probability rule defined on the graph. By employing multi-task learning, the whole model can be trained end-to-end. In this way, $HM^3$ can 1) efficiently use all the impression samples in the entire space to address the \emph{SSB} and \emph{DS} issues; 2) model both micro and macro behaviors in a hierarchical and complete manner by using a larger neural network compared with $ESM^2$ \cite{wen2020entire}; and 3) leverage the abundant supervisory labels from micro behaviors for learning better feature embeddings and facilitating the modeling of macro behaviors, which are more sparse compared with micro behaviors.

The main contributions of this paper are summarized as follows:

$\bullet$ To the best of our knowledge, we are the first to explicitly model both micro and macro behaviors in a unified framework for efficient and accurate \emph{CVR} prediction.

$\bullet$ We propose a novel deep neural recommendation model named $HM^{3}$, which can utilize all the impression samples and harvest the abundant labeled micro and macro post-click behaviors to efficiently mitigate the \emph{SSB} and \emph{DS} issues via multi-task learning.


$\bullet$ We conduct extensive experiments on both real-world offline dataset and online scenario and confirm the superiority of $HM^{3}$ over representative state-of-the-art methods.


\section{The Proposed Approach}
\label{sec:method}
\subsection{Motivation and Novel User Behavior Graph}
\label{subsec:motivation}

In ${ESM^2}$ \cite{wen2020entire}, specific purchase-related post-click behaviors are incorporated into the user sequential behavior graph, based on which a unified multi-task learning framework is proposed to harvest the abundant labeled data from these post-click behaviors. In this way, ${ESM^2}$ outperforms \emph{ESMM} \cite{ma2018entire}, which neglects post-click behaviors. Nevertheless, they both neglect abundant post-click micro behaviors, $e.g.$, clicks on certain components of items like comments or pictures, that are related to the macro behaviors and final purchase target. As discussed in the previous section, these micro behaviors are also useful for recommendation \cite{zhou2018micro,gu2020hierarchical,bao2020gmcm}. However, how to define their relationship with those macro behaviors as well as embody both micro and macro behaviors into a unified framework for \emph{CVR} prediction still remains unexplored. 

\begin{table}[ht]
\caption{Detailed behaviors in the D-Mi set and D-Ma set.}
\begin{tabular}{p{0.8cm}p{7cm}}
\hline
Type & Behaviors \\ \hline
D-Mi & clicking item's pictures, checking the Q\&A details of an item, chatting with sellers, reading an item's comments, clicking an item's carting control button\\
D-Ma & putting an item into the shopping cart or wish list \\ \hline
\end{tabular}
\label{table:mm_behaviors}
\end{table}


For one thing, several elaborately defined micro behaviors have explicit and deterministic labels which are more abundant compared with macro behaviors as illustrated in Figure~\ref{fig:HM3_motivation}(a). For another, both micro behaviors and macro behaviors belong to post-click behaviors. Therefore, it is promising to leverage the abundant labeled data from them to learn better feature representations and predict more accurate $CVR$. Following $ESM^{2}$ \cite{wen2020entire}, we construct a complete user sequential behavior graph to hierarchically represent micro behaviors and macro behaviors as one-hop and two-hop post-click nodes. Specifically, two nodes named Deterministic Micro set (\emph{D-Mi}) and Deterministic Macro set (\emph{D-Ma}) are defined to merge several predefined specific purchase-related post-click micro behaviors and macro behaviors, as listed in Table~\ref{table:mm_behaviors}. These two nodes have two properties: 1) they are highly relevant to respective following nodes and purchase behaviors; and 2) they have abundant labels derived from users' feedback, $e.g.$, 1 for taking any kinds of predefined micro (macro) behaviors and 0 for none. In order to account for different kinds of post-click behaviors completely, we also add two nodes named Other Micro set (\emph{O-Mi}) and Other Macro set (\emph{O-Ma}) to deal with other post-click behaviors except \emph{D-Mi} and \emph{D-Ma}, respectively. Then, we hierarchically insert the four defined nodes between click and purchase, \emph{D-Mi} and \emph{O-Mi} first, followed by \emph{D-Ma} and \emph{O-Ma}. In this way, the conventional behavior path ``impression$\to$click$\to$purchase'' is transformed into a new user sequential behavior graph as depicted as in Figure~\ref{fig:HM3_motivation}(b).

\subsection{Probability Decomposition on the Graph}
\label{subsec:probabilityDecompositionCVR}

Referring to Figure~\ref{fig:HM3_motivation}(b), given an item $x_i$, we define six probability variables $y_{1i}$, $y_{2i}$, $y_{3i}$, $y_{4i}$, $y_{5i}$ and $y_{6i}$ corresponding to explicit sub-paths in the graph, $i.e.$, ``impression$\to$click'', ``click$\to$\emph{D-Mi}'', ``\emph{D-Mi}$\to$\emph{D-Ma}'', ``\emph{D-Ma}$\to$purchase'', ``\emph{O-Mi}$\to$\emph{D-Ma}'' and ``\emph{O-Ma}$\to$purchase'', respectively. Next, we define four auxiliary tasks over the entire space denoted as $p_i^{ctr}$, $p_i^{D-Mi}$, $p_i^{D-Ma}$, and $p_i^{ctcvr}$, representing the probabilities of certain sub-paths , $i.e.$, ``impression$\to$click'', ``impression$\to$\emph{D-Mi}'', ``impression$\to$\emph{D-Ma}'', and ``impression$\to$purchase'', respectively. According to the conditional probability rule on the graph, the probabilities of these auxiliary targets can be derived from the aforementioned six probability variables as follows:
\begin{equation}
    p_{i}^{ctr}=y_{i1},
\label{eq:pctr}
\end{equation}
\begin{equation}
    p_{i}^{D-{Mi}}=y_{i1}*y_{i2},
\label{eq:pDMi}
\end{equation}
\begin{equation}
    p_{i}^{D-{Ma}}=y_{i1}*\left( y_{i2}*y_{i3}+(1-y_{i2})*y_{i5} \right),
\label{eq:DMa}
\end{equation}
\begin{equation}
\begin{aligned}
  p_{i}^{ctcvr} &= y_{i1} y_{i4} \left( y_{i2}y_{i3}+(1-y_{i2})y_{i5} \right) + \\
    & \quad y_{i1} y_{i6} \left( y_{i2}(1-y_{i3})+(1-y_{i2})(1-y_{i5}) \right) \\
    &= p_{i}^{ctr}*p_{i}^{cvr}.
 \end{aligned}
\label{eq:pctcvr}
\end{equation}
Here, $p_{i}^{cvr}$ denotes the probability of conversion rate of a clicked item $x_{i}$, which can be derived from the six probability variables.


\subsection{Multi-task Learning based on DNNs}
\label{subsec:MTL}
\begin{figure}
  \centering
  \includegraphics[width=\linewidth]{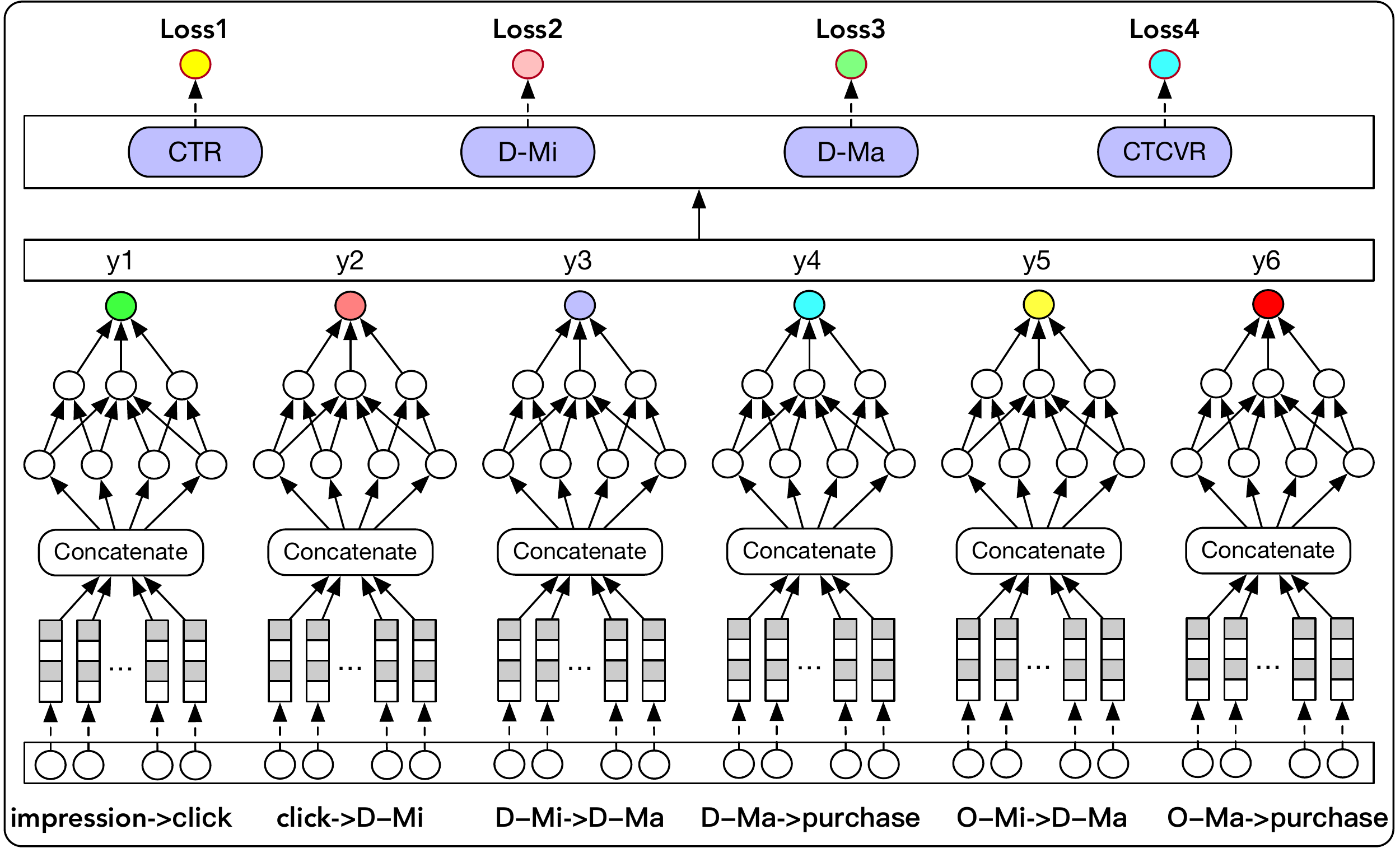}
  \caption{ 
  The diagram of the $HM^3$. It has six sub-networks to predict the probabilities of corresponding explicit sub-paths, which are composed sequentially to calculate the final CVR and other targets of four related auxiliary tasks.
  }
  \label{fig:HM_model}
\end{figure}

How to model the above six probability variables as well as those four auxiliary tasks simultaneously? An intuitive way is to embody them into a unified multi-task learning framework. To this end, we propose a novel deep neural recommendation model $HM^3$ by hierarchically modeling both micro and macro behaviors for \emph{CVR} prediction. The network structure of $HM^3$ is diagrammed in Figure~\ref{fig:HM_model}. Firstly, we devise six parallel sub-networks after a shared feature embedding module (FEM) to predict the probability variables. Since FEM makes up the majority of the parameters, sharing the FEM makes the whole network lightweight. In addition, FEM can receive abundant back-propagated supervisory signals to mitigate the \emph{DS} issue and learn good feature embeddings.


Then, based on the predicted conditional probabilities $y_{1i}\sim y_{6i}$ from the corresponding sub-networks, the probabilities of four auxiliary targets can be derived sequentially according to Eq.~\eqref{eq:pctr} $\sim$ Eq.~\eqref{eq:pctcvr}. Our model can not only efficiently estimate six probability variables in parallel but also do not require extra parameters for deriving $CVR$ and auxiliary targets, which are very suitable for online deployment and responding users' requests in a low latency. Finally, based on the supervisory labels from click behaviors, micro (macro) behaviors, and purchase behaviors corresponding to the four binary classification tasks, we can define the final training objective as a combination of four cross-entropy losses. Once the model being trained, we can derive $p_{i}^{cvr}$ from Eq.~\eqref{eq:pctcvr} accordingly. It is noteworthy that the \emph{SSB} issue is well addressed due to the fact that all the auxiliary tasks are modeled over the entire space. Besides, due to the hierarchically modeling over the user sequential behavior graph (referring to Figure~\ref{fig:HM3_motivation}(b) and Figure~\ref{fig:HM_model}), abundant supervisory signals from micro behaviors can better supervise the learning of macro behaviors, coordinately towards the final purchase target .


\vspace{-0.3cm}
\begin{table}
\caption{Statistics of three training sets of different volumes.}
\resizebox{\linewidth}{!}{%
\begin{tabular}{lllllllll}
\hline
Dataset & \#user & \#item & \#impression & \#click & \#D-Mi & \#D-Ma & \#purchase \\ \hline
SR-S & 32M & 50M & 4.9B & 146M & 36M & 19M & 5M \\
SR-M & 68M & 76M & 14.8B & 434M & 102M & 58M & 16M \\ 
SR-L & 107M & 98M & 31.7B & 925M & 235M & 122M & 32M \\ \hline
\end{tabular}
}
\label{table:stats}
\end{table}

\section{Experiments}
\label{sec:experiment}

\subsection{Experiment settings}
\label{subsec:settings}
\subsubsection{Dataset preparation}
\label{subsubsec:dataset}
We collected the consecutive logs between 2020-09-16 and 2020-09-30 from the \emph{Shopping Recommendation} (\emph{SR}) module of our recommender system \footnote{To the best of our knowledge, there are no available large-scale public datasets containing both micro and macro behaviors to benchmark our method.}. We used the last day data as the test set. To investigate the impact of data volume, we prepared three training sets: \emph{SR-S}, \emph{SR-M}, and \emph{SR-L}, containing data from 2020-09-29, 2020-09-23 to 2020-09-29, 2020-09-16 to 2020-09-29, respectively. Table~\ref{table:stats} summarizes their statistics.

\subsubsection{Competitors}
\label{subsubsec:sotaMethods}
1) $\boldsymbol{BASE}$~\cite{hinton2006fast}: a deep neural network model which uses the same input features as $HM^3$. The network structure and hyper parameters are identical to each single branch of $HM^3$, such as \emph{ReLU} for activation function, \emph{Adam} for optimizer, 0.0005 for learning rate, etc. It predicts the probabilities $p^{ctr}$ or $p^{cvr}$ using samples from ``impression$\to$click'' or ``click$\to$purchase'', respectively. 2) $\boldsymbol{ESMM}$~\cite{ma2018entire}: it predicts $p^{ctr}$ and $p^{cvr}$ over the entire space without using the purchase-related macro post-click behaviors. 3) $\boldsymbol{ESM^2}$~\cite{wen2020entire}: it models purchase-related post-click macro behaviors in a unified multi-task learning framework, while neglecting micro behaviors. 4) $\boldsymbol{GMCM}$~ \cite{bao2020gmcm}: it represents user micro behaviors as a graph and employs graph convolutional networks to model their interaction relationships towards the final purchase target. 5) $\boldsymbol{ESM^2+Mi}$: it uses the same network structure as $ESM^2$~\cite{wen2020entire}. The only difference is that it uses post-click behaviors from both macro behaviors as well as micro behaviors. 
6) $\boldsymbol{HM^3-R}$: it uses the same network structure as $HM^3$ while only reversing the order of micro and macro behaviors in the graph, $i.e.$, D-Ma (O-Ma) is in front of D-Mi (O-Mi).


\subsubsection{Evaluation Metrics}
\label{subsubsec:metrics}
We adopt the widely used Area Under Curve (\emph{AUC}) \cite{zhou2018deep, zhu2017optimized} as the evaluation metric, which indicates the ranking performance. Specifically, we report the \emph{AUC} scores for the CVR prediction task and CTCVR prediction task. The latter task measures the probability of purchase given impression. 

\subsection{Main results}
\label{subsec:mainResults}

\begin{table}[htbp]
\caption{The AUC scores of all methods for CVR. }
\begin{tabular}{c|c|c|c}
\hline
 & \multicolumn{1}{c|}{SR-S} & \multicolumn{1}{c|}{SR-M} & \multicolumn{1}{c}{SR-L} \\
\hline
Method & CVR AUC &CVR AUC &CVR AUC \\
\hline
BASE&0.84117   &0.84583    &0.84932 \\
\hline
ESMM&0.84297   &0.84601    &0.85098 \\
\hline
$ESM^2$&0.84493  &0.84794    &0.85213\\   
\hline
GMCM&0.84624  &0.84928    &0.85405\\   
\hline 
$ESM^2+Mi$&0.84703  &0.85118    &0.85596\\ 
\hline
$HM^3-R$&0.84799  &0.85285    &0.85646\\ 
\hline
$\boldsymbol{HM^3}$&\textbf{0.84891}   &\textbf{0.85328}    &\textbf{0.85726}\\
\hline
\end{tabular}
\label{tab:all_method_cvr}
\end{table}

Table~\ref{tab:all_method_cvr} and Table~\ref{tab:all_method_ctcvr} summarize the respective AUC results of different methods for the CVR and CTCVR tasks. Taking the results of set \emph{SR-L} in Table~\ref{tab:all_method_cvr} as an example, it can be seen that \emph{ESMM} achieves a gain of 0.00166 over the \emph{BASE} model, because the \emph{SSB} and \emph{DS} issues could be mitigated in \emph{ESMM} by modeling the problem over the entire space. As for $ESM^2$, it uses more abundant purchase-related post-click behaviors to alleviate the sparsity issue of positive training samples and therefore achieving a gain of 0.00115 over \emph{ESMM}. However, $ESM^2$ ignores the abundant micro behaviors, which are also related to the purchase target as well as macro behaviors. By using these micro behaviors, GMCM outperforms $ESM^2$ by 0.00192 AUC. After modeling micro behaviors, $ESM^2 + Mi$ improves the performance of the vanilla $ESM^2$ by 0.00383 AUC, confirming the value of exploring micro behaviors. It also outperforms GMCM by 0.00191 AUC, indicating the benefit of modeling both kinds of behaviors together. Our $HM^3$ obtains the best results among all the methods. For example, it outperforms the \emph{BASE} model by 0.00794 AUC. Compared with $ESM^2 + Mi$ and $HM^3-R$, $HM^3$ also performs better, which demonstrates that hierarchically modeling the micro behaviors and the macro behaviors is more effective. Similar results can be found on other two datasets for the CVR task as well as for the CTCVR task. It is worth mentioning that the slightly better AUC score at the setting of a larger dataset volume means a significant increment in revenue for online deployment according to our practice.

\begin{table}[htbp]
\caption{The AUC scores of all methods for CTCVR. }
\begin{tabular}{c|c|c|c}
\hline
 & \multicolumn{1}{c|}{SR-A} & \multicolumn{1}{c|}{SR-B} & \multicolumn{1}{c}{SR-C} \\
\hline
Method & CTCVR AUC &CTCVR AUC &CTCVR AUC \\
\hline
BASE&0.84883   &0.85532   &0.85993 \\
\hline
ESMM&0.85051   &0.85652    &0.86192 \\
\hline
$ESM^2$&0.85148  &0.85804    &0.86296\\    
\hline
GMCM&0.85241  &0.85953    &0.86411\\   
\hline 
$ESM^2+Mi$&0.85402  &0.86098    &0.86596\\  
\hline
$HM^3-R$&0.85516  &0.86163    &0.86712\\  
\hline
$\boldsymbol{HM^3}$ &\textbf{0.85675}   &\textbf{0.86213}    &\textbf{0.86806}\\
\hline
\end{tabular}
\label{tab:all_method_ctcvr}
\end{table}

We also conducted online A/B test between different methods and the $BASE$ model in the \emph{SR} modules of our recommendation platform from 2020-10-08 to 2020-10-21. The results are summarized in Table~\ref{tab:online_ab}. $HM^3$ brings $\boldsymbol{8.27\%}$ CVR gain and $\boldsymbol{8.32\%}$ GMV (Gross Merchandise Volume) gain over the $BASE$ model, $i.e.$, larger margins than other methods, which are consistent with the offline evaluation results, indicating a significant business revenue growth.

\begin{table}[htbp]
  \caption{Results of Online A/B test.}
  \label{tab:data}
  \begin{tabular}{lll}
    \toprule
  Method & CVR Gain & GMV Gain \\
    \midrule
  BASE & 0\% & 0\%  \\
  ESMM & +2.76\% & +3.02\%  \\
  $ESM^2$ & +4.84\% & +5.11\% \\
  $\boldsymbol{HM^3}$ & +8.27\% & +8.32\% \\
  \bottomrule
  \hline
\end{tabular}
\label{tab:online_ab}
\end{table}

\vspace{-0.6cm}
\section{Conclusion}
\label{sec:Conclusion}

We propose a novel deep model $HM^3$ for CVR prediction. It hierarchically models micro and macro post-click behaviors in a unified multi-task learning framework according to a new user sequential behavior graph. $HM^3$ efficiently mitigates the \emph{SSB} and \emph{DS} issues by leveraging abundant supervisory signals from the behaviors and all impression samples over the entire space. Experiments on offline datasets and online test show $HM^{3}$ outperforms state-of-the-art models. In the future, we intend to investigate the impact of modeling post-click behaviors at a finer granularity.

\balance
\bibliographystyle{ACM-Reference-Format}
\bibliography{2020SIGIR_RS}

\end{document}